\documentclass[10pt]{article}
\usepackage[preprint]{tmlr}
\usepackage{amsmath}
\usepackage{bm}
\usepackage{graphicx}
\usepackage{algorithm}
\usepackage{algorithmic}

\title{ILP-BO: Integer Linear Programming-Based Black-Box Optimization}

\author{\name Hyakka Nakada \email hyakka\_nakada@r.recruit.co.jp \\
      \addr Recruit Co., Ltd.
      \AND
      \name Shu Tanaka \email shu.tanaka@keio.jp\\
      \addr Graduate School of Science and Technology, Keio University \\
      Department of Applied Physics and Physico-Informatics, Keio University \\
      Keio University Sustainable Quantum Artificial Intelligence Center (KSQAIC), Keio University \\
      Human Biology-Microbiome-Quantum Research Center (WPI-Bio2Q), Keio University}

\begin{document}

\maketitle

\begin{abstract}
Black-box Optimization (BO) is a powerful framework for optimizing expensive objective functions or unknown functions with a limited number of evaluations. A central step of standard BO such as Bayesian optimization is the optimization of a surrogate-based acquisition criterion, which is commonly performed using nonlinear optimization or heuristic search. 
Therefore, conventional black-box optimization generally does not guarantee global optimality in candidate selection. 
In this study, we propose Integer Linear Programming-based Black-box Optimization (ILP-BO), a quasi-Bayesian optimization framework that transforms kernel-based surrogate optimization over discrete domains into an Integer Linear Programming (ILP) problem. The key idea is to represent nonlinear kernel functions exactly on finite discrete distance levels by introducing binary one-hot auxiliary variables. This transformation converts the nonlinear surrogate into a linear objective with linear constraints and binary variables. 
To incorporate exploration while preserving the linear structure, we further introduce a Hamming-distance margin that excludes neighborhoods around previously observed points. We derive the proposed formulation for several standard kernels and obtain an analytical upper bound on the Hamming-distance threshold based on the measure in the binary search space. 
The resulting candidate-selection problem can be solved by integer programming solvers with certificates of optimality. Thus, our methodology has the potential to serve as a highly transparent black-box optimization framework.
Experiments on synthetic and discrete optimization benchmarks show that ILP-BO achieves competitive optimization performance compared with practical Bayesian optimization methods.
\end{abstract}

\section{Introduction}
\label{sec:introduction}
Black-Box Optimization (BO) of unknown functions arises in a wide variety of practical problems, including hyperparameter optimization, drug discovery, protein design, and hardware optimization~\citep{jo:98}. In many such applications, evaluating the black-box objective is expensive, and therefore optimization methods are required to identify good solutions with as few function evaluations as possible.
Bayesian optimization is a powerful framework for such expensive BO problems~\citep{ga:23,sh:16,fr:18}. A surrogate model is trained from previously observed data, and an acquisition criterion is then constructed to determine a promising next point. 
Here, the optimum of the acquisition function is searched.
After evaluating the black-box function at the selected point, the observation data is added to the dataset and the surrogate model is updated. This cycle is repeated until the available evaluation budget is exhausted.

Gaussian Processes (GPs) are widely used as surrogate models in Bayesian optimization~\citep{ra:05,sn:12}. 
The resulting acquisition functions are generally nonlinear in the candidate variables and are often optimized using local or heuristic search~\citep{am:23}. Such procedures do not generally guarantee the global optimum of the acquisition criterion, particularly in high-dimensional or discrete search spaces.
An alternative is to reformulate candidate selection for mathematical optimization solvers that can certify global optimality.
Recent studies have investigated approximations of kernel functions and acquisition criteria using piecewise-linear or quadratic representations, enabling the use of mathematical optimization solvers. These methods improve the reliability of the candidate selection process but may result in quadratic or otherwise complicated optimization problems, which leads to the difficulty of optimization in the case of high-dimensional problem size.
Another drawback is that the approximation of the acquisition function is inevitable.

\begin{figure}[tpb!]
\centering
\includegraphics[width=10cm]{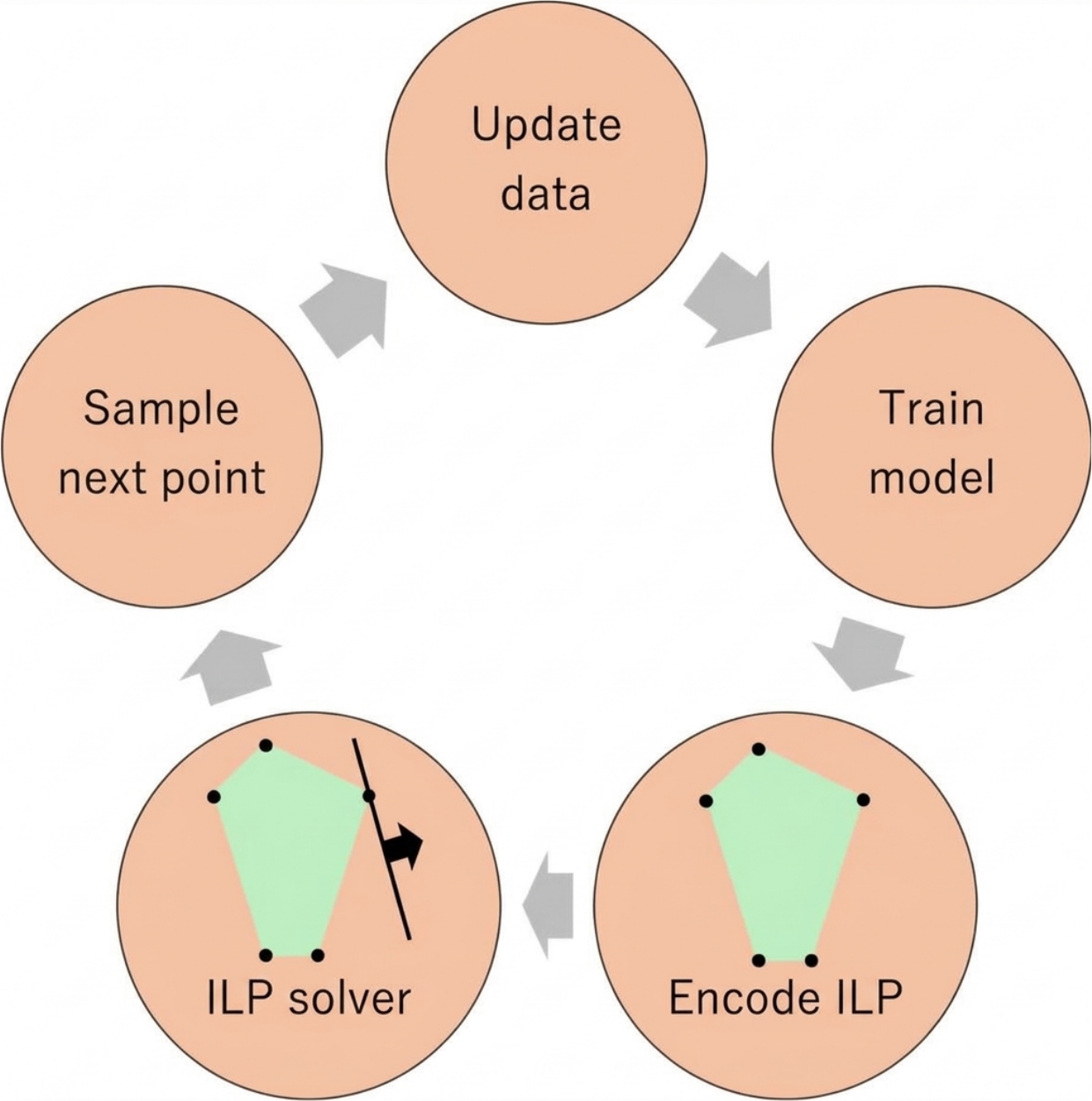}
\caption{Concept of ILP-BO. In addition to the conventional BO process, the trained surrogate model is encoded into an ILP problem. The next point is then predicted by maximizing or minimizing this problem with an ILP solver.}
\label{fig:ILP-BO}
\end{figure}

We propose Integer Linear Programming-based Black-box Optimization (ILP-BO), a framework for candidate selection in binary black-box optimization, as illustrated in Fig~\ref{fig:ILP-BO}.
We utilize integer linear programming solvers to obtain the maximum point of the surrogate model. Recently, while GP is highly nonlinear in its original form, the surrogate model can be expressed in a low dimensional form by introducing appropriate additional variables.
In this study, we show that modifying this theory can encode the objective function into a linear function with linear constraints. 

Our construction applies to kernel surrogates whose distance or similarity arguments take finitely many values and admit affine representations in the binary decision variables.
In other words, the high degree of degeneracy in a kernel function is leveraged.
By introducing binary one-hot variables, the surrogate optimization problem can therefore be formulated as an Integer Linear Program (ILP).

A challenge in this formulation is how to define exploration. Direct optimization of the surrogate mean tends to emphasize exploitation. In conventional GP-based BO, exploration is often incorporated through predictive uncertainty, but explicit covariance terms generally introduce nonlinear dependencies that are difficult to keep an ILP formulation. We therefore introduce a Hamming distance margin that excludes neighborhoods around previously observed points. This mechanism provides a simple form of exploration while preserving the integer-linear structure, as shown later.
As a result, both the exploitation and exploration can be handled by an ILP.

Our main contributions are as follows:
\begin{itemize}
\item We establish an exact linearization framework for a class of kernel-based surrogate functions on discrete domains using binary one-hot auxiliary variables.
\item We formulate surrogate optimization as a 0--1 integer linear program and show that the formulation can accommodate several commonly used kernels.
\item We introduce a Hamming-distance margin as an exploration mechanism that preserves linearity and derive an analytical upper bound for the margin based on the coverage of the discrete search space.
\item We experimentally evaluate ILP-BO on several discrete optimization benchmarks and compare its optimization behavior with conventional BO methods.
\end{itemize}

\section{Related Work}
\label{sec:related}
Standard BO methods such as Bayesian optimization construct a surrogate model from observed data and optimize an acquisition criterion derived from the surrogate model, to maximize unknown true objective function $G(\bm{x})$.
A surrogate model $F(\bm{x})$ is trained on initial data $(\bm{x}_k,G(\bm{x}_k))$ for $k=0,1,\ldots,K_0-1$ to construct an acquisition function.
Here, $K_0$ is the number of initial data.
Then, the next promising point $\bm{x}_{\text{next}}$ is predicted by maximizing the acquisition function. 
After the new observation $(\bm{x}_{\text{next}},G(\bm{x}_{\text{next}}))$ is fed back to the dataset, the surrogate model is re-trained. 

Gradient-based methods such as L-BFGS-B~\citep{zh:97} and stochastic gradient ascent~\citep{ki:15} are widely used for acquisition optimization. Although effective in many settings, these methods generally identify local optima and do not provide a guarantee of global optimality~\citep{wi:18}. Sampling-based methods such as Nelder--Mead~\citep{ne:65} provide an alternative, but exhaustive or globally optimal search becomes difficult as the dimensionality of the search space increases. Few works have focused on achieving global optimality.

For a notable study on the optimality at each iteration, there is a report that kernel functions are approximated by piecewise lines to improve BO performance~\citep{xi:25}. In the study, the approximate acquisition function becomes quadratic polynomial with quadratic constraints. This problem can be solved by quadratic Mixed Integer Programming (MIP)~\citep{le:11}, improving the optimality.
However, such solvers have difficulty of solving problems of a larger size than linear programming solvers. Particularly, in the case of discrete black-box optimization problems, finding the optimal solution is fundamentally difficult because quadratically constrained quadratic programs are known to be $\mathcal{NP}$--hard.
In addition, this method requires the approximation of acquisition function.

While their method is specialized for a continuous domain and gradient descent methods were partially utilized, we focus on a discrete domain, and we have proved that a maximization process in BO can be effectively reduced into an ILP with linear constraints. Our formulation is not only approximation-free, but also gradient descent-method free.
Previous works~\citep{xi:25, ki:20, mi:25, ta:26} have explored approximating the acquisition function using quadratic forms.
ILP-BO provides the linearity task designed for BO.

A wide variety of solvers has been developed for an ILP, enabling the exact solution of remarkably large-scale problems using not only commercial but also non-commercial solvers~\citep{ga:03, hu:18, su:24}. Recently, several reports showed that such problems can be efficiently solved by quantum-inspired tensor networks~\citep{ha:22, lo:23, na1:25}.
Thus, the superiority in global optimality over quadratic programming solvers will further intensify as problem scale grows.

\section{Integer Linear Programming-Based Black-Box Optimization}
\label{sec:ILP-BO}

We first derive the discrete linearization of kernel-based surrogate functions. We then introduce an exploration mechanism based on Hamming distance and derive a feasibility bound for its threshold. Finally, we summarize the resulting optimization procedure.

\subsection{Discrete Linearization of Kernel Regressions}
\label{subsec:discrete_linearization}
Consider a surrogate function of the form
\begin{equation}
F(\bm{x})
=\sum_{k=0}^{K-1} c_k f(q_k(\bm{x})),
\label{eq:1}
\end{equation}
where $q_k(\bm{x}):\bm{x} \to R^1$ is a scalar map function, $f$ is a nonlinear function, and $c_k$ is a real-valued coefficient. 
Generally, kernel regressions can be written in Eq.~\eqref{eq:1} form by adopting $f$ as a kernel function.
Especially, a natural number $K$ means the number of training data in the case of kernel regressions.

Hereinafter, suppose that $q_k(\bm{x})$ takes one of $L$ discrete values
$d_{k,0},d_{k,1},\ldots,d_{k,L-1}$.
We introduce binary auxiliary variables $s_{k,l}\in \{0,1\}$, according to the previous study~\citep{na2:25}.
Then, the nonlinear function can be represented exactly as
\begin{align}
f(q_k(\bm{x}))
=&\sum_{l=0}^{L-1}
f(d_{k,l})s_{k,l},
\notag\\
\mathrm{s.t.}\quad
\sum_{l=0}^{L-1}s_{k,l}=&1, \ 
\sum_{l=0}^{L-1}
d_{k,l}s_{k,l}
=q_k(\bm{x}).
\label{eq:2}
\end{align}
The first constraint in Eq.~\eqref{eq:2} imposes the one-hot condition to determine a single value out of levels, while the second ensures consistency between $q_k(\bm{x})$ and the selected level.

Substituting Eq.~\eqref{eq:2} into Eq.~\eqref{eq:1}, we obtain
\begin{align}
\max_{\bm{x}}F(\bm{x})
=\max_{\bm{x},\bm{s}}
\quad&
\sum_{k=0}^{K-1}
\sum_{l=0}^{L-1}
c_k f(d_{k,l})s_{k,l}
\notag\\
\mathrm{s.t.}\quad
&
\sum_{l=0}^{L-1}
d_{k,l}s_{k,l}
=q_k(\bm{x}),
\quad k=0,\ldots,K-1,
\notag\\
&
\sum_{l=0}^{L-1}s_{k,l}=1,
\quad k=0,\ldots,K-1.
\label{eq:3}
\end{align}
While the above focuses on maximizing the function $F$, minimization can also be handled by modifying the operators.

When $q_k(\bm{x})$ is linear in the original binary variables $\bm{x}$, Eq.~\eqref{eq:3} is a 0--1 ILP problem. Therefore, the nonlinear optimization of the surrogate function can be transformed into an integer-linear optimization problem without approximating the kernel function.

Eq.~\eqref{eq:3} requires $KL$ additional binary variables $\bm{s}$ in addition to the original variables $\bm{x}$. Here, the maximization operator for $\bm{s}$ is additionally adopted for convenience to unify operators with respect to $\bm{x}$.
The constraints of Eq.~\eqref{eq:3} have trivial feasible solutions, such as $\bm{x} = \bm{x}_k$ and $s_{k,0} = 1$ (i.e., corresponding to the observed data themselves).

\subsection{Binary Domains and Hamming Distance}

We consider a binary search space, $\bm{x}\in \{0,1\}^{D}$.
Here, $D$ is the dimension of $\bm{x}$, in other words the problem size. 
For the RBF kernel $f=e^{-q_k(\bm{x})/2\sigma^2}$, the squared Euclidean distance is used,
\begin{equation}
q_k(\bm{x})=|\bm{x}-\bm{x}_k|_2^2=\sum_{i=0}^{D-1}
\left(
(1-2x_{k,i})x_i+x_{k,i}
\right).
\label{eq:4}
\end{equation}
Here, $\sigma^2$ means a variance hyper parameter.
For binary variables, this quantity is equal to the Hamming distance.
Therefore, $q_k(\bm{x})\in\{0,1,\ldots,D\}$, and the number of possible levels is $L=D+1$.
The ILP form of the RBF kernel is given by
\begin{equation}
f(d_{k,l})
=\exp
\left(
-\frac{d_{k,l}}{2\sigma^2}
\right),
\quad
d_{k,l}=l.
\label{eq:5}
\end{equation}
The total number of binary auxiliary variables is $K(D+1)$ before the exploration margin is introduced.

The same construction can be applied to other kernels whose arguments can be expressed by finite-valued functions of the discrete input. For example, for the Mat\'{e}rn kernel $\frac{2^{1-\nu}}{\Gamma(\nu)}(\sqrt{2\nu}|\bm{x}-\bm{x}_k|/\sigma)^{\nu}K_{\nu}(\sqrt{2\nu}|\bm{x}-\bm{x}_k|/\sigma)$,
\begin{equation}
f(d_{k,l})
=\frac{2^{1-\nu}}{\Gamma(\nu)}
\left(
\frac{\sqrt{2\nu d_{k,l}}}{\sigma}
\right)^{\nu}
K_{\nu}
\left(
\frac{\sqrt{2\nu d_{k,l}}}{\sigma}
\right),
\quad
d_{k,l}=l.
\label{eq:6}
\end{equation}
Here, $\nu$ is a hyper parameter that adjusts differentiability.
$\Gamma(\nu)$ and $K_{\nu}$ are the Gamma function and the modified Bessel function of the second kind, respectively.

For the rational quadratic kernel $(1+|\bm{x}-\bm{x}_k|^2)^{-\gamma}$,
\begin{equation}
f(d_{k,l})
=(1+d_{k,l})^{-\gamma},
\quad
d_{k,l}=l.
\label{eq:7}
\end{equation}
Here, $\gamma$ is a positive hyperparameter.

For the Ornstein--Uhlenbeck kernel $e^{-|\bm{x}-\bm{x}_k|/\sigma}$,
\begin{equation}
f(d_{k,l})
=\exp
\left(
-\frac{\sqrt{d_{k,l}}}{\sigma}
\right),
\quad
d_{k,l}=l.
\label{eq:8}
\end{equation}

For the white kernel $\delta_{\bm{x},\bm{x}_k}$,
\begin{equation}
f(d_{k,l})
=\begin{cases}
1,&l=0,\\
0,&l>0.
\end{cases}
\quad
d_{k,l}=l.
\label{eq:9}
\end{equation}

Moreover, the above encoding can be applied to other kernels based on non-Hamming distance. For the polynomial kernel $(\bm{x}^T \bm{x}_k+c)^{\gamma}$,
\begin{equation}
f(d_{k,l})=(d_{k,l}+c)^{\gamma},
\quad
d_{k,l}=l.
\label{eq:10}
\end{equation}
Here, $c$ is a constant.
$q_k(\bm{x})=\bm{x}^T \bm{x}_k=0,1,\ldots,D$, and the total number of levels is still $L=D+1$.
In this case, $q_k(\bm{x})$ represents an inner product rather than a Hamming distance. The same integer-linearization principle still applies because the possible values remain finite.

\subsection{Combined Kernels}

The formulation can be extended to a sum of multiple kernel functions,
\begin{equation}
F(\bm{x})=
\sum_{k=0}^{K-1}
\left(
c_k f(q_k(\bm{x}))
+
c'_k f'(q'_k(\bm{x}))
\right).
\label{eq:11}
\end{equation}

By introducing independent one-hot variables for the two kernels, the optimization can be formulated as
\begin{align}
\max_{\bm{x},\bm{s},\bm{s}'}
\quad&
\sum_{k=0}^{K-1}
\left\{
\sum_{l=0}^{L-1}
c_k f(d_{k,l})s_{k,l}
+
\sum_{l=0}^{L'-1}
c'_k f'(d'_{k,l})s'_{k,l}
\right\}
\notag\\
\mathrm{s.t.}\quad
&
\sum_{l=0}^{L-1}
d_{k,l}s_{k,l}
=q_k(\bm{x}),
\quad k=0,\ldots,K-1,
\notag\\
&
\sum_{l=0}^{L-1}s_{k,l}=1,
\quad k=0,\ldots,K-1,
\notag\\
&
\sum_{l=0}^{L'-1}
d'_{k,l}s'_{k,l}
=q'_k(\bm{x}),
\quad k=0,\ldots,K-1,
\notag\\
&
\sum_{l=0}^{L'-1}s'_{k,l}=1,
\quad k=0,\ldots,K-1,
\notag\\
&
s_{k,l},s'_{k,l}\in\{0,1\}.
\label{eq:12}
\end{align}
Thus, composite surrogate functions can also be optimized within an integer-linear formulation.

\subsection{Exploration Through Hamming-Distance Margin}
\begin{figure}[t]
\centering
\includegraphics[width=12cm]{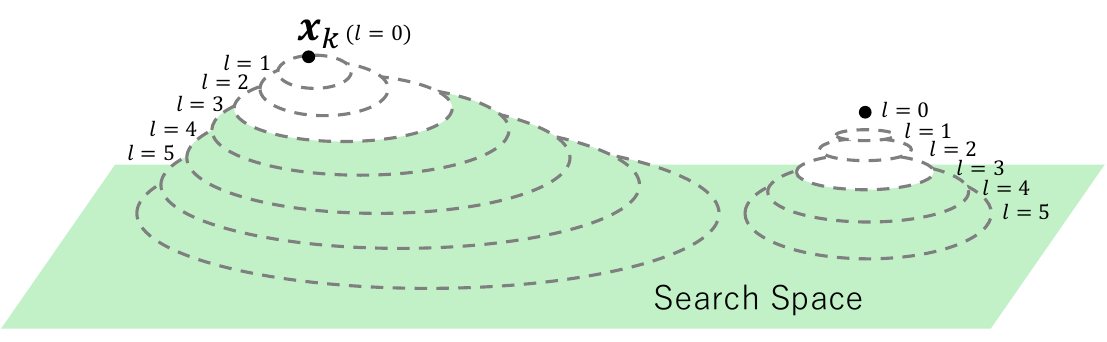}
\caption{Schematic picture of the search space in ILP-BO. 
Topographic surface of surrogate model is shown as dashed mountains.
To encode original problem into an ILP problem, additional variables $s_{k,l}$ are introduced. Their index $l$ describes a Hamming distance from training data point $\bm{x}_k$. Search space for $l\geq 3$ is shown in green area. Thus, outside neighborhoods of $\bm{x}_k$, next points are determined by ILP-BO.
For simplicity, slices at each value of $l$ are depicted parallel to the $\bm{x}$ plane. In addition, this plane is drawn simply in a 2D map. Although kernels such as RBF tend to have training data points located on summits as depicted, this does not always coincide in general.}
\label{fig:search_space}
\end{figure}

\begin{figure}[t]
\centering
\includegraphics[width=10cm]{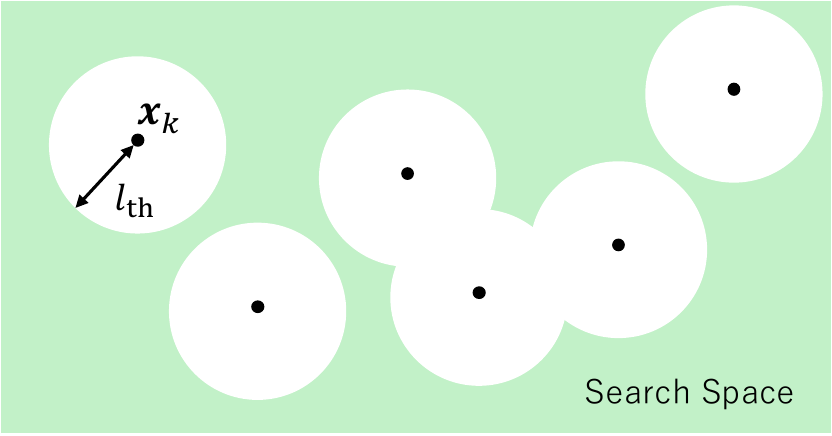}
\caption{Projection of Fig.~\ref{fig:search_space} onto $\bm{x}$ plane. For simplicity, $l_{\text{th}}$-neighborhoods around training data points are depicted as disk areas with the points centered. Finite $l_{\text{th}}$ can eliminate candidates similar to existing training data.}
\label{fig:search_space_2d}
\end{figure}

Although Eq.~\eqref{eq:3} provides an exact optimization formulation for the surrogate model, directly optimizing it emphasizes exploitation and does not explicitly reflect predictive uncertainty.
We instead exploit the fact that, for the distance-based kernels considered above, $q_k(\bm{x})$ can be interpreted as a dissimilarity from the observed point $\bm{x}_k$.

By adjusting the value of $l$, new data different from the existing observed data can be explored, as shown in Fig.~\ref{fig:search_space}.
For the binary domain, $q_k(\bm{x})=l$ is the Hamming distance. In particular, 
\begin{equation}
q_k(\bm{x})=0
\quad\Longleftrightarrow\quad
\bm{x}=\bm{x}_k.
\label{eq:13}
\end{equation}
Thus, when $s_{k,0}$ is hot, the candidate $\bm{x}$ corresponds to the observed point $\bm{x}_k$.
If $s_{k,1}$ is hot, $\bm{x}$ will shift from $\bm{x}_k$ to another state with a Hamming distance of $1$ away.

Selecting a larger value of $q_k(\bm{x})= l$ therefore forces the candidate to be further from an observed point.
We introduce a threshold $l_{\mathrm{th}}$ and restrict the candidate search to states with the Hamming distance at least $l_{\mathrm{th}}$ from every observed point, as shown in Fig.~\ref{fig:search_space_2d}. The resulting optimization problem is
\begin{align}
\max_{\bm{x},\bm{s}}
\quad&
\sum_{k=0}^{K-1}
\sum_{l=l_{\mathrm{th}}}^{L-1}
c_k f(d_{k,l})s_{k,l}
\notag\\
\mathrm{s.t.}\quad
&
\sum_{l=l_{\mathrm{th}}}^{L-1}
d_{k,l}s_{k,l}
=q_k(\bm{x}),
\quad k=0,\ldots,K-1,
\notag\\
&
\sum_{l=l_{\mathrm{th}}}^{L-1}s_{k,l}=1,
\quad k=0,\ldots,K-1,
\notag\\
&
s_{k,l}\in\{0,1\}.
\label{eq:14}
\end{align}
Here, $0\le l_{\mathrm{th}}\le L-1$.
A larger $l_{\mathrm{th}}$ excludes a larger neighborhood around previously observed points and therefore places greater emphasis on exploration. When $l_{\mathrm{th}}=0$, previously observed points remain eligible. When $l_{\mathrm{th}}>0$, the candidates are predicted from the unexplored area.

An additional consequence is that the number of auxiliary binary variables becomes $K(L-l_{\mathrm{th}})$, which decreases as the threshold $l_{\mathrm{th}}$ increases. Conventional Bayesian optimization relies on predictive variance for exploration, which makes the optimization problem higher-order and more complex~\citep{sr:10}. In contrast, the proposed method utilizes a margin strategy for exploration, which actually reduces the number of variables and thereby simplifies the optimization problem.

\subsection{Feasibility of Hamming-Distance Margin}
\begin{figure}[t]
\centering
\includegraphics[width=10cm]{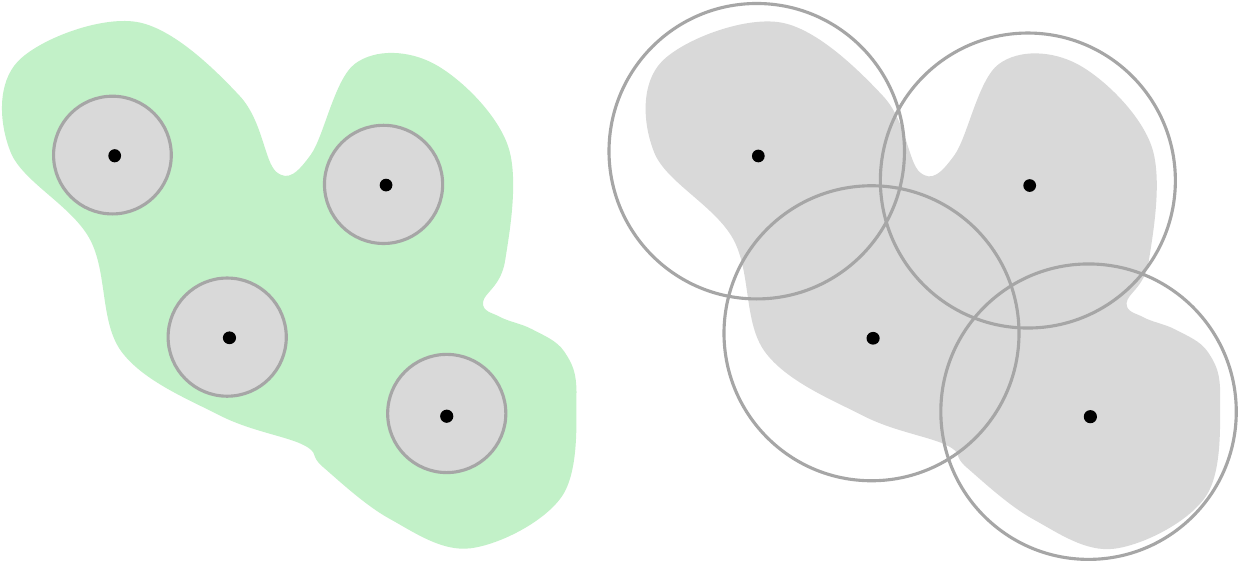}
\caption{Covarage of entire space of $\bm{x}$ with $l_{\text{th}}$-neighborhoods. To avoid infeasibility, value of $l_{\text{th}}$ must be smaller than that of lower bound to cover entire space, leading to Eq.~\eqref{eq:14}.}
\label{fig:covered_space}
\end{figure}

Note that excessive values of $l_{\text{th}}$ may lead to infeasibility because the entire feasible search space is eliminated. 
As explained in Subsection~\ref{subsec:discrete_linearization}, setting $l_{\text{th}}=0$ guarantees the existence of feasible solutions. However, for example, if $l_{\text{th}}=D$, the entire space of $\bm{x}$ is trivially excluded from the search space.
In Fig.~\ref{fig:search_space_2d}, one can imagine that the search space vanishes with a large $l_{\text{th}}$.
In other words, when the entire space of $\bm{x}$ is covered with the white space of $l<l_{\text{th}}$, Eq.~\eqref{eq:14} becomes infeasible.
If the entire space of $\bm{x}$ is not covered, solutions are searched for the complemented subspace.

We therefore derive an approximate upper bound based on the coverage of the binary hypercube, as shown in Fig.~\ref{fig:covered_space}.
The upper bound of $l_{\text{th}}$ can be deduced based on a density function for the Hamming distances.
The density of states with the Hamming distance of $h$ follows the binomial distribution $h\sim\mathcal{B}(D,1/2)$~\citep{ca:20}.
According to the central limit theorem, this distribution is asymptotically approximated by Gaussian distribution,
\begin{equation}
h\sim\mathcal{N}\left(\frac{D}{2}, \frac{D}{4} \right),
\label{eq:15}
\end{equation}
for sufficiently large $D$.

We assume that there are $K$ training data points and these points are evenly scattered throughout the space, which is the pessimistic scenario of $l_{\text{th}}$ estimation.
In this case, as shown in Fig.~\ref{fig:covered_space}, to ensure that the entire space should not be covered with $l_{\text{th}}$-neighborhoods of the data points,
\begin{equation}
1
\ge
K
\int_{-\infty}^{l_{\mathrm{th}}}
\sqrt{\frac{2}{\pi D}}
\exp
\left(
-\frac{2(t-D/2)^2}{D}
\right)
dt.
\label{eq:16}
\end{equation}
Conceptually, Eq.~\eqref{eq:16} shares the fundamental geometric principle of the Gilbert-Varshamov (GV) bound for sphere covering, from the viewpoint of the coding theory~\citep{ma:77}.
While traditional coding-theoretic formulations rely on discrete binomial sums, our formulation adopts the Gaussian approximation~\eqref{eq:15} to yield an explicit closed-form bound.

Let $\Phi$ denote the cumulative distribution function of the standard Gaussian distribution $\mathcal{N}(0,1)$. Then an approximate upper bound is
\begin{equation}
l_{\mathrm{th}}
\le
l_{\mathrm{up}}(K,D)
=\max
\left\{
\min
\left(
\frac{D}{2}
+
\frac{\sqrt{D}}{2}
\Phi^{-1}
\left(
\frac{1}{K}
\right),
D
\right),
0
\right\}.
\label{eq:17}
\end{equation}

From Eq.~\eqref{eq:17}, for example, $l_{\text{up}}(K,D)= D/2$ is obtained for $K=2$.
This is reasonable because if we consider two poles at $(0,0,\ldots,0)$ and $(1,1,\ldots,1)$, the minimum Hamming distance between another point and these points is always lower than $D/2$. 
For $K=1000$, the approximate value of the upper bound for $D=10$ is $0.11<1$.
This result implies that finite $l_{\text{th}}$ (i.e., $\geq 1$) cannot be allocated in the case of $(K,D)=(1000,10)$, which is reasonable because the entire space $2^{10}=1024$ is almost covered by the $1000$ existing data points.
In addition, the upper bound decreases as the number of observations increases. For $K=10$, the approximate values of the upper bound for $D=10,20,30,40$ are $2.97$, $7.13$, $11.49$, and $15.95$, respectively. For $K=1000$, the corresponding values are approximately $0.11$, $3.09$, $6.54$, and $10.23$.

We used sufficiently small values of $l_{\text{th}}$ according to Eq.~\eqref{eq:17} in the later experiments, and no infeasible solutions were confirmed to appear.

There is one drawback in our formalism. 
The number of additional variables is $K(L-l_{\text{th}})$, which increases linearly with the number of training data $K$.
Thus, the calculation time on an ILP solver increases as the optimization steps progress.
Although such solvers have developed remarkably with respect to computational performance, for practical use, it is desirable to use them in short-step BO.
For long-step BO, to keep the number of training data $K$ constant, one can consider removing unimportant data, such as the point $\bm{x}$ with the smallest $G(\bm{x})$ value in the pre-update data.

\subsection{Optimization Procedure}

\let\AND\undefined
\begin{algorithm}[t]
\caption{ILP-BO for discrete black-box optimization}
\label{alg:ilpbo}
\begin{algorithmic}[1]
\REQUIRE Kernel type $f$, initial sample size $K_0$, initial threshold $l_{\mathrm{th}}(1)$, decay ratios
\ENSURE Best observed solution
    \STATE Sample initial data ($K_0$ data)
    \FOR{$t =1,2,\cdots,T$}
      \STATE Train kernel regressors with existing data to form the surrogate model (Eq.~\eqref{eq:1})
      \STATE Set Hamming distance threshold $l_{\text{th}}(t)$ by Eq.~\eqref{eq:19} or Eq.~\eqref{eq:20}
      \STATE Create the linear programming problem of Eq.~\eqref{eq:14}
      \STATE Solve the problem with an ILP solver to obtain next point $\bm{x}_{\text{next},t}$
      \IF{$\bm{x}_{\text{next},t}$ is feasible} 
        \STATE Update data by evaluating $G(\bm{x}_{\text{next},t})$, and update the number of data $K_t=K_0+t$
      \ELSE
        \STATE continue
      \ENDIF
    \ENDFOR
    \STATE Choose the best point from the $K_T$ data points
\end{algorithmic}
\end{algorithm}

The optimization flow of ILP-BO is summarized in Algorithm~\ref{alg:ilpbo}.
First, initial data is sampled. Generally, this is performed uniformly at random.
Then, the following process is iterated until the preset number of trials $T$ is completed.
At the $t$-th step ($t=1,\ldots,T$), a surrogate model $F$ is trained with the existing data.
The model is a kernel regression~\eqref{eq:1} with the preset kernel $f$.
Here, the available types of kernel are the RBF, rational quadratic, Mat\'{e}rn, white, Ornstein–Uhlenbeck, or a combination thereof.
This is because they have the characteristic that $l$ reflects a Hamming distance.
Then, an ILP problem of Eq.~\eqref{eq:14} is created based on the trained model $F$.
Note that the margined version of Eq.~\eqref{eq:12} is created alternatively if a combination of the above kernels is adopted as $f$.
Next, the problem is solved by ILP solvers to obtain the next point $\bm{x}_{\text{next},t}$.
If $\bm{x}_{\text{next},t}$ is feasible, the data are updated with the new observation $(\bm{x}_{\text{next},t},G(\bm{x}_{\text{next},t}))$.
Otherwise, the data are not updated.
Finally, the best point is selected by referring to $\bm{x}$ with the highest $G(\bm{x})$ in the data.

In the above flow, the basic and natural schedule for Hamming distance threshold is exponential decay,
\begin{equation}
l_{\mathrm{th}}(t;r)
=\min
\left\{
1+
\left(
l_{\mathrm{th}}(1)-1
\right)r^{t-1},
l_{\mathrm{up}}(K_t,D)
\right\}.
\label{eq:18}
\end{equation}
Here, 
$l_{\text{th}}(t;r)$ denotes the Hamming distance threshold at step $t$, and $0\leq r \leq 1$ is a decay scale, that is a geometric ratio for exponential decrease. $K_t$ is the total number of data at step $t$.
Eq.~\eqref{eq:18} is designed so as to converge to $1$ and be bounded above by the upper limit Eq.~\eqref{eq:17}.

The decay scale $r$ plays a critical role in balancing exploration and exploitation over the optimization process. 
A larger $r$ maintains a wider Hamming-distance margin $l_{\mathrm{th}}$ for a longer duration, promoting global exploration, whereas a smaller $r$ rapidly shrinks the margin, prioritizing local exploitation around previously observed points.

With a single scale $r$, a monotonous balance between exploitation and exploration may lead to performance variations.
Therefore, in this study, we adopt not only the single-scale schedule
\begin{equation}
l_{\mathrm{th}}(t)
=l_{\mathrm{th}}(t;r), \ \text{for} \ t=1,2,\ldots,T,
\label{eq:19}
\end{equation}
but also the following multiple-scale schedule, 
\begin{equation}
l_{\mathrm{th}}(t)
=\begin{cases}
l_{\mathrm{th}}(t;r_1), & t=1,4,7,\ldots,\left\lfloor \frac{T}{3} \right\rfloor -2\\
l_{\mathrm{th}}(t;r_2), & t=2,5,8,\ldots, \left\lfloor \frac{T}{3} \right\rfloor-1\\
l_{\mathrm{th}}(t;r_3), & t=3,6,9,\ldots, \left\lfloor \frac{T}{3} \right\rfloor.
\end{cases}
\label{eq:20}
\end{equation}
Here, $\left\lfloor x \right\rfloor$ is a floor function.
Equation~\eqref{eq:20} has a conservative scale $r_1$, a fair scale $r_2$, and an aggressive scale $r_3$ ($r_1>r_2>r_3$).
These decay scales were determined in Subsection~\ref{subsec:decay_scale}.

\section{Numerical Experiments}

We evaluate ILP-BO on five representative discrete optimization benchmarks:
\begin{itemize}
\item Sphere function~\citep{sh:99},
\item Max-cut problem~\citep{go:95},
\item Number partitioning problem~\citep{bo:01},
\item Ising model~\citep{ci:87},
\item Low Autocorrelation Binary Sequences (LABS) problem~\citep{bo:67}.
\end{itemize}

Because the proposed method is a significantly new type of solver, the sphere function, which is one of the most basic benchmarks, is used for principle verification.
The Max-Cut problem, the Number Partitioning problem, and the Ising model are well-known standard benchmarks for discrete optimization, which are equivalent to a quadratic unconstrained binary optimization problem~\citep{pa:22}.
Finally, as a practical problem, the LABS problem is considered. 
The LABS problem is a higher-order problem, which is classically intractable even for moderately sized instances~\citep{pa:16}. This problem has real-world applications in communications engineering, in which low autocorrelation sequences are utilized to design radar pulses.

As \texttt{ILP-BO}, we adopt two variations (\texttt{Single} and \texttt{Multi}) based on the schedule types~\eqref{eq:19} and \eqref{eq:20}.
To rigorously evaluate their performance, we compare \texttt{ILP-BO} against a comprehensive set of baselines:
\begin{itemize}
\item \texttt{Random Search}: Uniform random sampling across the binary domain.
\item \texttt{BO-UCB}: GP-based Bayesian optimization utilizing the upper confidence bound acquisition function in \texttt{BayesianOptimization}~\citep{no:14}. In particular, we employ a method adapted for discrete variables~\citep{ga:20}.
\item \texttt{BOCS}: Bayesian Optimization for Combinatorial Structures~\citep{ba:18}, using semidefinite programming relaxation combined with Goemans-Williamson randomized rounding on horseshoe-regularized second-order surrogates.
\item \texttt{Optuna GPS}: GP-based Bayesian optimization sampler in \texttt{Optuna}~\citep{ak:19}.
\item \texttt{Optuna TPES}: Tree-structured parzen estimator sampler~\citep{be:11} for non-GP discrete search in \texttt{Optuna}.

\end{itemize}

\subsection{Common Experimental Settings}
In this study, we adopt the Mat\'{e}rn kernel~\eqref{eq:6} with $\nu=2.5$, which is also used by default in \texttt{Optuna}. 
Its hyperparameter, the length scale $\sigma$, is optimized through maximum likelihood estimation.
This optimization and model training are performed with \texttt{Scikit-learn}~\citep{pe:11}.
Here, the length scale bounds and the initial length scale are set to $(1, 2D)$ and $D$, respectively, for \texttt{Matern} class in \texttt{Scikit-learn}.
Then, a GP model is constructed using \texttt{GaussianProcessRegressor} with \texttt{n\_restarts\_optimizer} $=20$.
After training the model with the \texttt{fit} method, using the estimated coefficients $c_k$ and hyperparameter $\sigma$, the ILP problem~\eqref{eq:14} is created.
As an ILP solver, \texttt{HiGHS}, one of the fastest non-commercial solvers for MIP~\citep{hu:18}, is adopted.

For all experiments, the number of initial data is set to $K_0=5$.
This data is generated randomly, and we ensure that any two points do not coincide.
For each problem, the problem size is varied between $D=20,30,40$.
We perform $10$ independent runs with differently generated initial data, and evaluate the performance based on average scores.
All algorithms share the exact same objective function instances and identical initial data ($K_0=5$) for each independent experimental trial.

The initial threshold is determined by $l_{\text{th}}(1)=\text{int}(D/4)$. The preset number of trials $T$, that is the total number of iterations, is set to $50$.
Here, modest values were set to avoid the computational intractability of exact mathematical optimization.
Verification over longer trials is a future issue. In other words, assessing scalability on larger-scale problems using high-performance commercial solvers~\citep{cp:09, gu:24} represents an important direction.
Nevertheless, the choice of $T=50$ reflects expensive real-world applications where evaluation budgets are severely constrained, making early-stage sample efficiency far more critical than long-horizon convergence.

In all conducted experiments, all candidate selections produced by \texttt{ILP-BO} remained strictly feasible, and the ILP solver confirmed global optimality of the acquisition surrogate at every iteration within the designated solver termination tolerances.

For all GP-based baselines (\texttt{BO-UCB} and \texttt{Optuna GPS}), $\nu=2.5$ Mat\'{e}rn kernel is used in the same manner as \texttt{ILP-BO}.
In addition, the hyperparameter optimization is commonly conducted with $20$ restarts ($n_{\mathrm{restarts\_optimizer}}=20$) and target normalization ($\mathrm{normalize\_y}=\mathrm{True}$) in these methods, to guarantee a fair comparison.

\subsection{Benchmark Problems}
We state the mathematical formulations of the discrete benchmark problems evaluated in our experiments.

\subsubsection{Sphere Function}
For binary decision variables $\bm{x}\in\{0,1\}^D$, the discrete Sphere function is defined as
\begin{equation}
F_{\mathrm{sphere}}(\bm{x})
=\sum_{i=0}^{D-1}x_i^2=\sum_{i=0}^{D-1}x_i.
\label{eq:21}
\end{equation}
The global minimum objective value is $F_{\mathrm{sphere}}(\bm{x}^*) = 0$, uniquely achieved at the origin $\bm{x}^*=(0,\ldots,0)$.

\subsubsection{Max-Cut Problem}
The Max-Cut problem aims to partition the vertex set of an undirected graph $G=(V,E)$ to maximize the total weight of edges straddling the two partitions. Let $w_{ij}$ denote the edge weight between vertices $i$ and $j$, sampled uniformly $w_{ij} \sim U(1.0, 5.0)$ with an edge density of $0.5$. Using spin variables $\sigma_i\in\{-1,1\}$, the objective to be minimized is expressed as
\begin{equation}
F_{\mathrm{maxcut}}(\bm{\sigma})
=-\sum_{i<j}
w_{ij}\frac{1-\sigma_i\sigma_j}{2}.
\label{eq:22}
\end{equation}
The spin variables are linearly mapped to binary space via $\sigma_i=1-2x_i$.

\subsubsection{Number Partitioning Problem}
The Number Partitioning problem partitions a set of $D$ positive integers $a_0, a_1, \ldots, a_{D-1}$ (sampled uniformly $a_i \sim U(1, 100)$) into two subsets to minimize the discrepancy of their subset sums. Representing subset choices via spin variables $\sigma_i\in\{-1,1\}$, the objective to be minimized is the squared sum difference,
\begin{equation}
F_{\mathrm{NP}}(\bm{\sigma})
=\left(\sum_{i=0}^{D-1}a_i\sigma_i\right)^2.
\label{eq:23}
\end{equation}
Binary domain mapping follows $\sigma_i=1-2x_i$.

\subsubsection{Ising Model}
The Ising model evaluates energy minimization over a fully connected spin system governed by the Hamiltonian
\begin{equation}
F_{\mathrm{ising}}(\bm{\sigma})
=\sum_{0\le i<j\le D-1}
J_{ij}\sigma_i\sigma_j,
\label{eq:24}
\end{equation}
where interaction coefficients $J_{ij}$ are drawn independently from $\mathcal{N}(0,1)$. Binary domain mapping follows $\sigma_i=1-2x_i$.

\subsubsection{LABS Problem}
The LABS problem minimizes the sequence autocorrelation energy
\begin{equation}
E_{\mathrm{LABS}}(\bm{\sigma})
=\sum_{j=1}^{D-1}
\left(
\sum_{i=0}^{D-j-1}
\sigma_i\sigma_{i+j}
\right)^2.
\label{eq:25}
\end{equation}
The corresponding quality measure is the Merit Factor
\begin{equation}
F_{\mathrm{LABS}}
=\frac{D^2}{2E_{\mathrm{LABS}}},
\label{eq:26}
\end{equation}
which is maximized in our optimization trials. Binary domain mapping follows $\sigma_i=1-2x_i$.

\subsection{Decay Scale Dependency}
\label{subsec:decay_scale}
To investigate the empirical sensitivity and quantitative trade-offs of the decay scale $r$, we evaluated the performance of \texttt{ILP-BO} under various values of $r \in \{0.80, 0.85, 0.90, 0.95, 0.97, 0.99\}$ using the Ising Model benchmark ($D=20$). Table~\ref{tab:performance_metrics} summarizes the average score improvement from iteration $0$ to $50$, as well as the progress rates in the early stage ($0 \to 10$) and the late stage ($30 \to 50$), defined as the ratio of score improvement during the respective interval relative to the total improvement.

\begin{table}[htbp]
  \centering
  \caption{Comparison of performance and progress rates in early and late stages for various decay scales. The $D=20$ Ising Model benchmark was used. The scores were averaged over $10$ initial-dataset samples.}
  \label{tab:performance_metrics}
  \begin{tabular}{ccccc}
    \hline
    Decay Scale $r$ & Total Improvement & Early Progress Rate & Late Progress Rate \\
    &  (0$\to$50) & (0$\to$10) & (30$\to$50) \\
    \hline
    $0.99$ &  28.09 & 56.2\% & 22.5\% \\
    $0.97$ &  29.41 & 53.6\% & 15.5\% \\
    $0.95$ &  30.63 & 52.4\% & 9.6\% \\
    $0.90$ &  31.07 & 59.3\% & 12.4\% \\
    $0.85$ &  30.64 & 60.4\% & 9.3\% \\
    $0.80$ &  30.46 & 64.1\% & 11.3\% \\
    \hline
  \end{tabular}
\end{table}
\begin{figure}[th!]
\centering
\includegraphics[width=16.5cm]{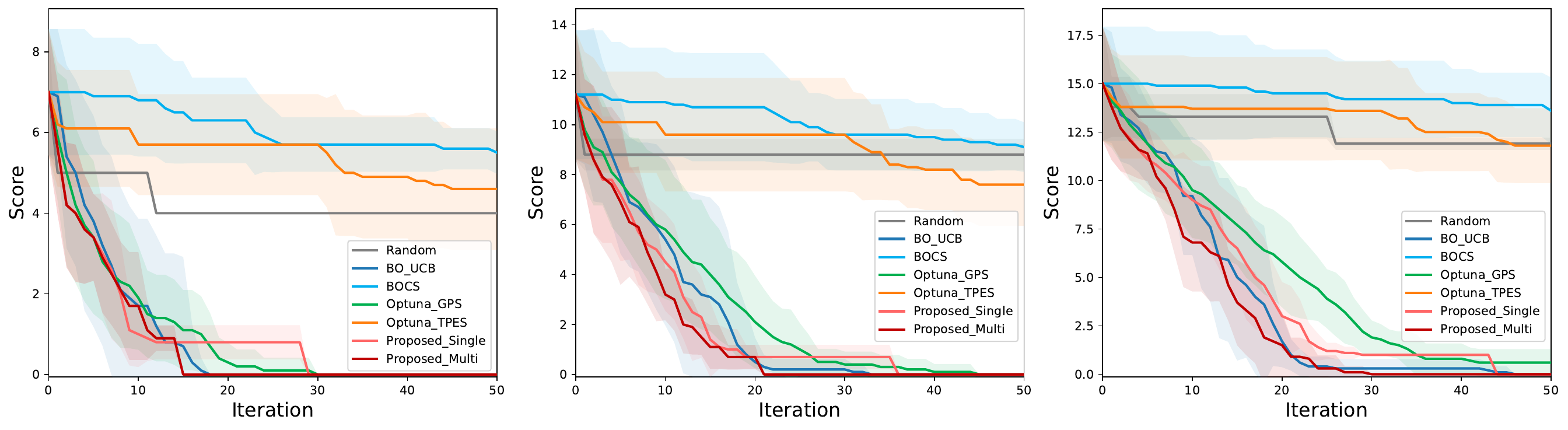}
\caption{Optimization performance on the Sphere benchmark for dimensions $D=20$ (left), $30$ (middle), and $40$ (right). Curves represent the mean scores over independent runs, with shaded regions depicting standard deviations. These lines shall guide the eye. Iterations are counted after initial data collection ($K_0=5$). Best observed objective values up to each iteration are shown (lower is preferred).
}
\label{fig:sphere_result}
\end{figure}

\begin{figure}[th!]
\centering
\includegraphics[width=16.5cm]{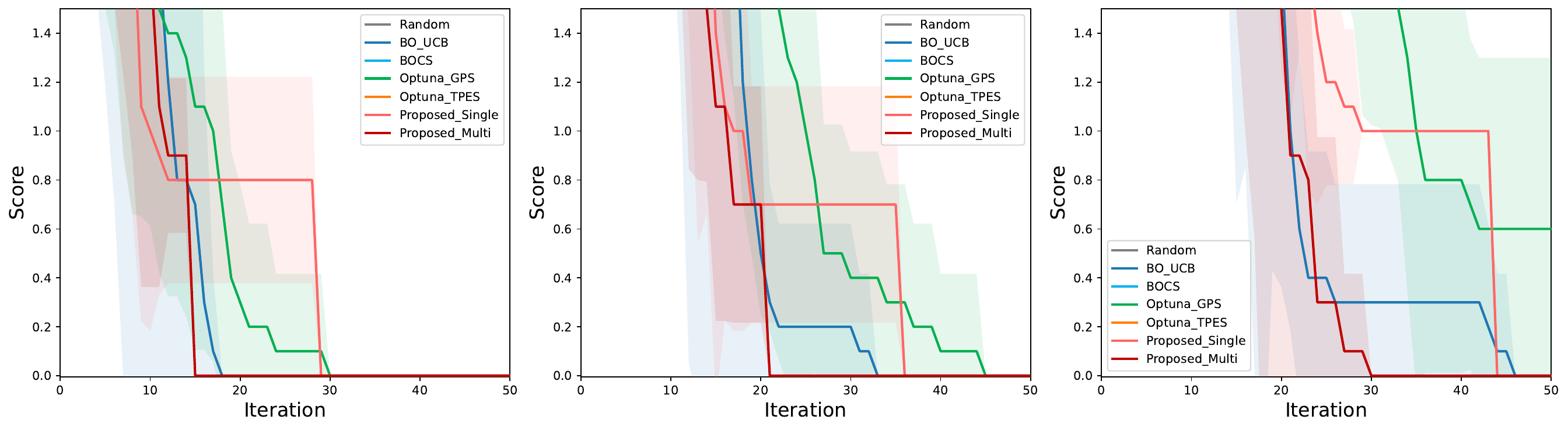}
\caption{Zoomed-in trajectory views corresponding to Fig.~\ref{fig:sphere_result}.}
\label{fig:sphere_zoom_result}
\end{figure}

As observed in Table~\ref{tab:performance_metrics}, there is no significant difference in overall improvement, indicating that the proposed method is generally robust in this benchmark.
However, a clear trade-off exists between early-stage acceleration and late-stage refinement. Smaller values of $r$, such as $r = 0.80$, achieved the highest early progress rate. However, this aggressive decay restricted broad space exploration, causing premature saturation and resulting in a lower total improvement of $30.46$. Conversely, larger values of $r$, such as $r = 0.99$, sustained high exploration capability into the late stages, yielding the maximum late progress rate of $22.5\%$. 
Nevertheless, delayed exploitation reduced overall sample efficiency within the fixed evaluation budget, leading to a smaller total improvement of $28.09$. Intermediate values of $r$ effectively achieved the optimal balance between these two extremes. 

Based on these observations, for our single-scale schedule in Eq.~\eqref{eq:19}, we adopt $r = 0.95$ as the default parameter. Furthermore, to overcome the inherent trade-off of any single fixed decay rate, we introduce the multiple-scale schedule in Eq.~\eqref{eq:20} with $r_1 = 0.97$, $r_2 = 0.95$, and $r_3 = 0.90$.

\subsection{Optimization Results}
\begin{figure}[tb!]
\centering
\includegraphics[width=16.5cm]{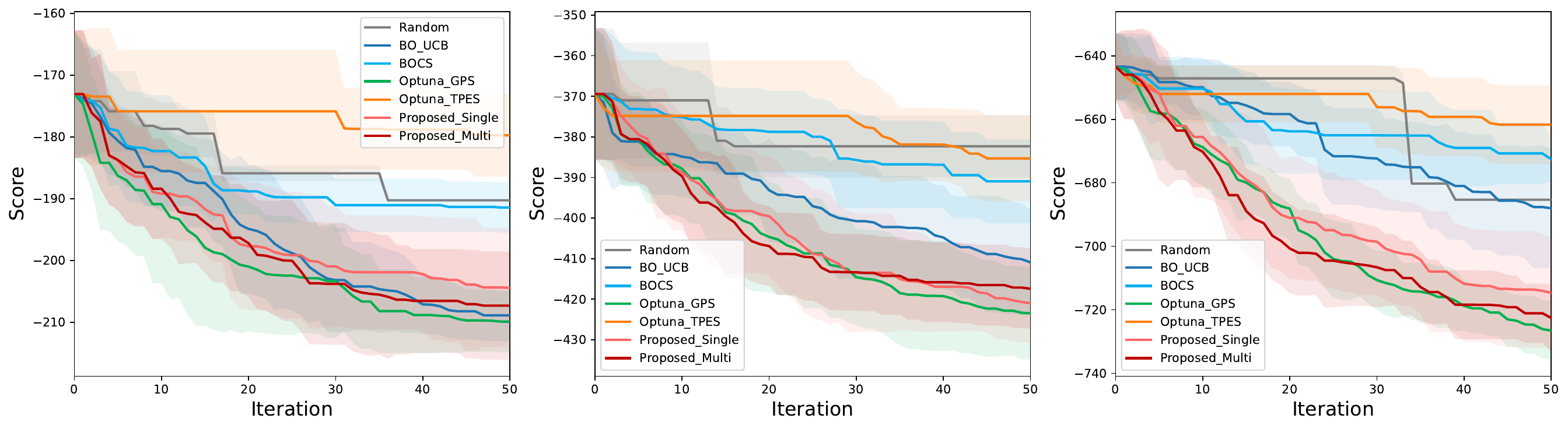}
\caption{Optimization performance on the Max-Cut problem ($D=20, 30, 40$). Best observed objective values up to each iteration are shown (lower is preferred).}
\label{fig:maxcut_result}
\end{figure}
\begin{figure}[tb!]
\centering
\includegraphics[width=16.5cm]{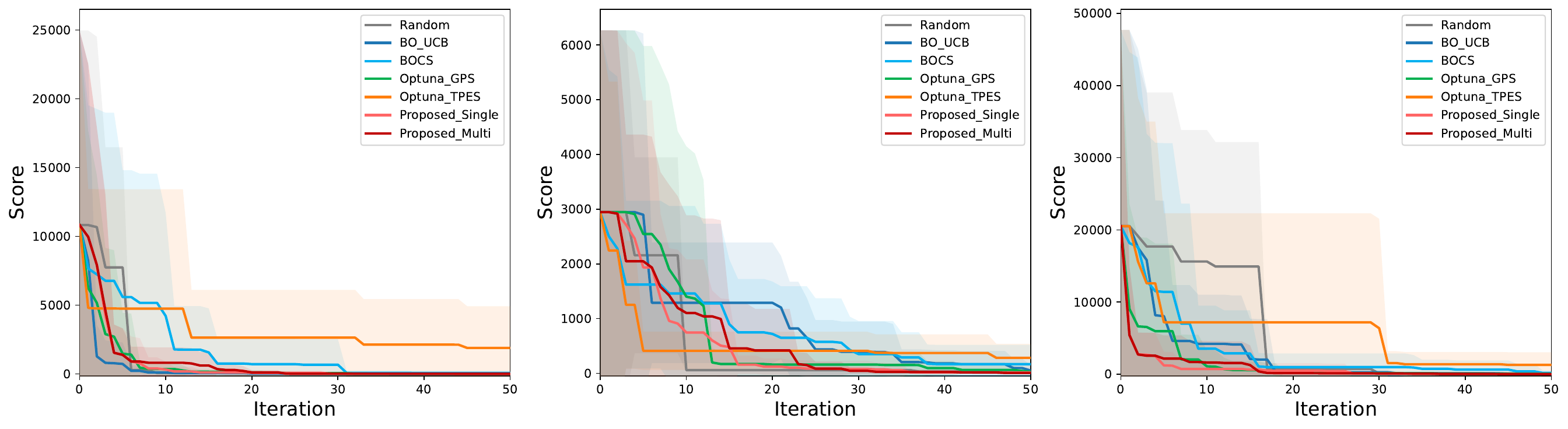}
\caption{Optimization performance on the Number Partitioning problem ($D=20, 30, 40$). Best observed objective values up to each iteration are shown (lower is preferred).}
\label{fig:number_result}
\end{figure}
\begin{figure}[tb!]
\centering
\includegraphics[width=16.5cm]{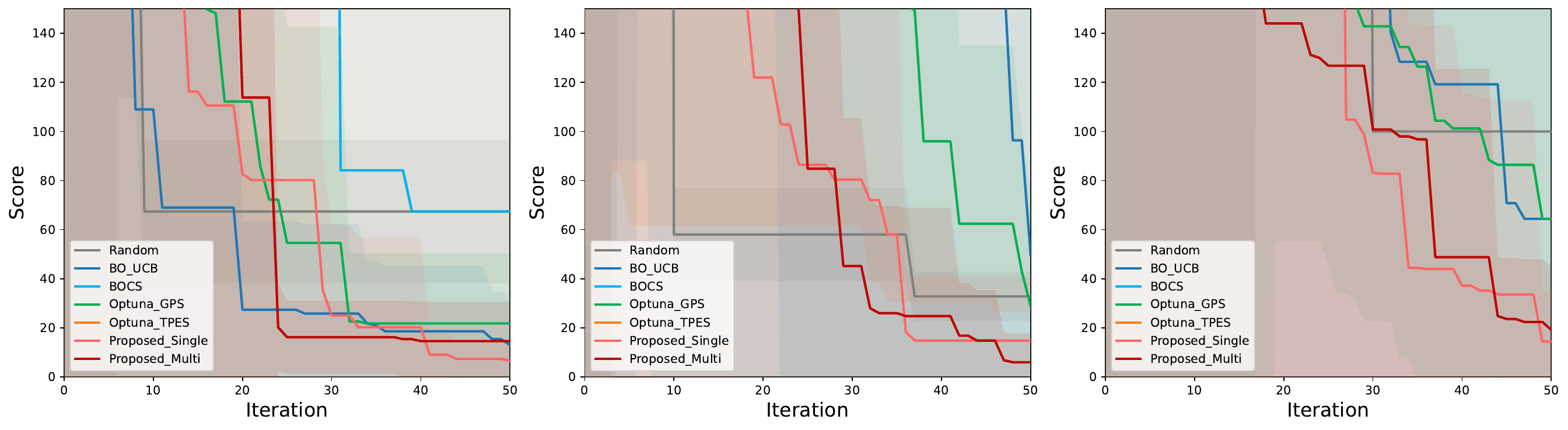}
\caption{Zoomed-in trajectory views corresponding to Fig.~\ref{fig:number_result}.}
\label{fig:number_zoom_result}
\end{figure}
\begin{figure}[tb!]
\centering
\includegraphics[width=16.5cm]{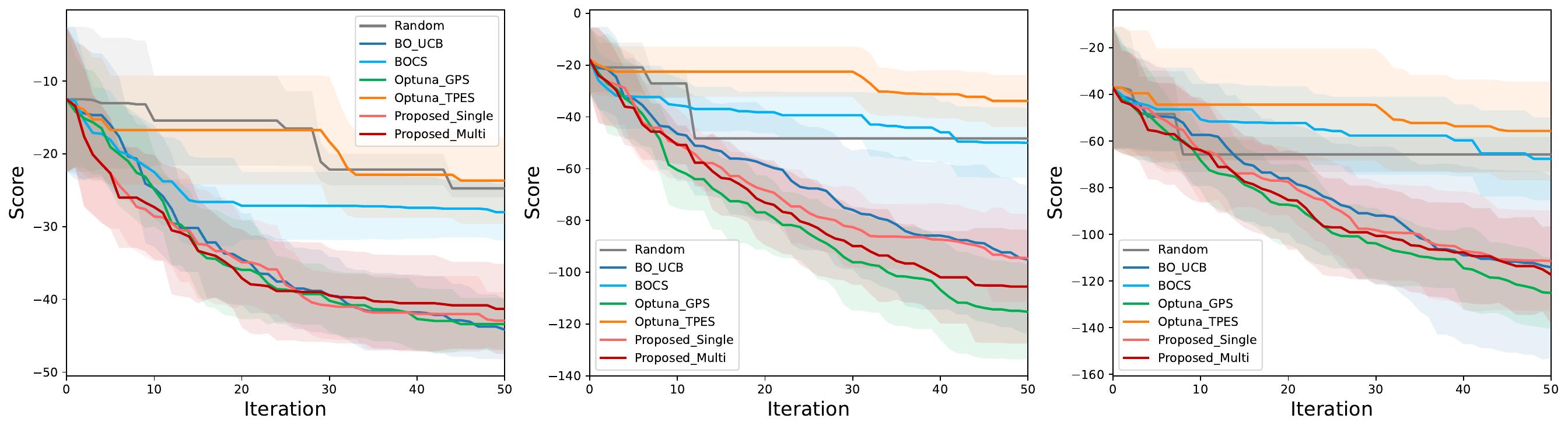}
\caption{Optimization performance on the Ising model ($D=20, 30, 40$). Best observed objective values up to each iteration are shown (lower is preferred).}
\label{fig:ising_result}
\end{figure}
\begin{figure}[tb!]
\centering
\includegraphics[width=16.5cm]{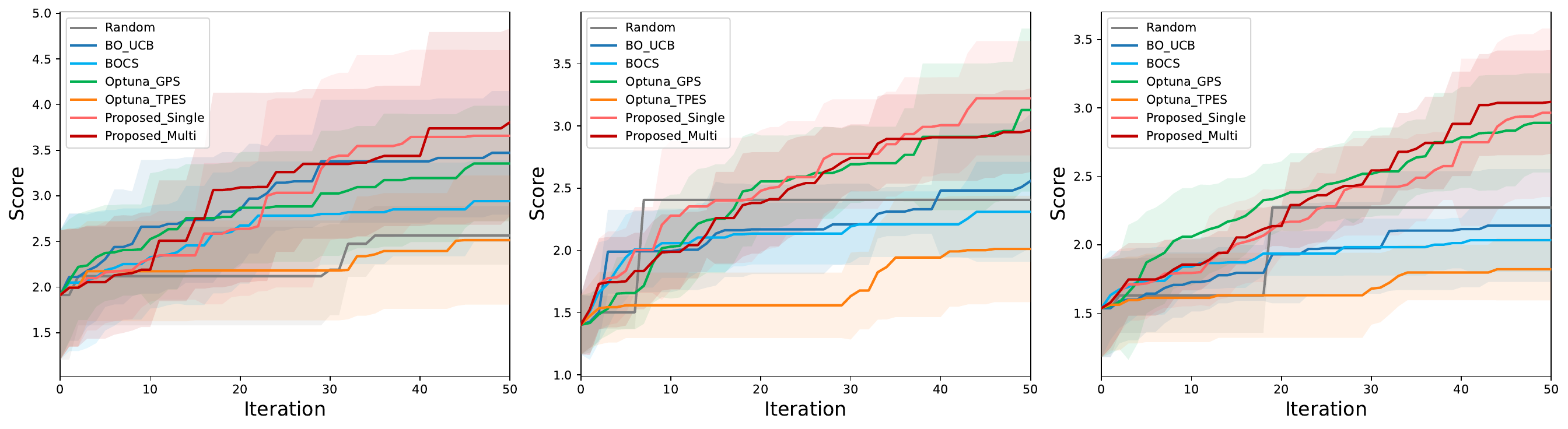}
\caption{Optimization performance on the LABS problem ($D=20, 30, 40$). Best observed objective values up to each iteration are shown (higher is preferred).}
\label{fig:labs_result}
\end{figure}

Figures~\ref{fig:sphere_result}--\ref{fig:labs_result} summarize the optimization trajectories on the five discrete benchmarks. The curves show the mean best-observed objective values, with the same initial observations shared across all methods. 
For \texttt{ILP-BO}, both the single-scale and multi-scale Hamming-distance schedules are evaluated.
Overall, the proposed method exhibits sample efficiency in the early stages of optimization, although the relative advantage depends on the benchmark and dimensionality.

On the Sphere benchmark (Figs.~\ref{fig:sphere_result} and
\ref{fig:sphere_zoom_result}), \texttt{ILP-BO} with the both proposed schedules rapidly reduced the objective toward the unique global optimum. Relatively, the multi-scale schedule generally reached the optimum in fewer evaluations. The advantage became particularly apparent in the higher-dimensional settings, where several baseline methods required substantially more evaluations to approach the optimum. Note that although \texttt{Optuna GPS} generally exhibited good performance on other problems, it showed poorer performance on the Sphere benchmark, the rate of improvement slowed down noticeably as the problem dimension increased.
Sometimes it failed to converge to the optimal solution.
On the other hand, the proposed methods and \texttt{BO-UCB} successfully found optimal solutions in all cases.

For the Max-Cut benchmark (Fig.~\ref{fig:maxcut_result}), \texttt{Optuna GPS} exhibited the best performance.
The proposed methods generally exhibited the second and third best performance. While \texttt{BO-UCB} yielded favorable results at the low dimension ($D=20$), its performance degraded as the dimensionality increased. For larger problem instances, the performance was almost comparable to that of Random Search.
The proposed method exhibited more robust performance compared to \texttt{BO-UCB}.

The Number Partitioning benchmark results (Figs.~\ref{fig:number_result} and \ref{fig:number_zoom_result}) provided one of the strongest empirical demonstrations of the proposed approach. \texttt{ILP-BO} reached the lowest scores.
Although some baselines continued to improve over the longer evaluation horizon, the early-stage advantage of \texttt{ILP-BO} remained pronounced. This behavior was important for expensive black-box optimization, where the number of evaluations rather than the asymptotic performance was often the primary practical constraint.


For the Ising model (Fig.~\ref{fig:ising_result}), \texttt{Optuna GPS} exhibited the lowest scores while \texttt{ILP-BO} was competitive to \texttt{BO-UCB}.

Finally, the LABS benchmark (Fig.~\ref{fig:labs_result}) demonstrated that the proposed framework was also effective for a discrete objective. \texttt{ILP-BO} achieved the highest Merit Factors throughout the optimization. 
While \texttt{BO-UCB} yielded favorable results at the low dimension ($D=20$), its performance degraded as the dimensionality increased. For larger problem instances, the performance degraded below that of \texttt{Random Search}.

Based on these results, the proposed method exhibited top-tier optimization performance compared to the baselines. On average, it discovered better scores than \texttt{BO-UCB} while achieving performance comparable to \texttt{Optuna GPS}. Furthermore, among the proposed variants, the multi-scale strategy tended to outperform the single-scale strategy.

\subsection{Global Optimum Attainment}
\begin{table}[tb!]
  \centering
  \caption{Success rate of reaching the exact optimum for Sphere and Number Partitioning problems}
  \label{tab:success_rate}
  \begin{tabular}{lcccccc}
    \hline
    & \multicolumn{3}{c}{Sphere} & \multicolumn{3}{c}{Number Partitioning} \\
    Problem Size & $D=20$ & $D=30$ & $D=40$ & $D=20$ & $D=30$ & $D=40$ \\
    \hline
    Random & 0.0 & 0.0 & 0.0 & 0.1 & 0.0 & 0.0 \\
    BO-UCB & \textbf{1.0} & \textbf{1.0} & \textbf{1.0} & 0.5 & 0.4 & \textbf{0.3} \\
    BOCS & 0.0 & 0.0 & 0.0 & \textbf{0.6} & 0.1 & 0.2 \\
    Optuna GPS & \textbf{1.0} & \textbf{1.0} & 0.5 & 0.5 & 0.4 & 0.1 \\
    Optuna TPES & 0.0 & 0.0 & 0.0 & 0.2 & 0.0 & 0.0 \\
    Proposed Single & \textbf{1.0} & \textbf{1.0} & \textbf{1.0} & 0.4 & \textbf{0.6} & 0.0 \\
    Proposed Multi & \textbf{1.0} & \textbf{1.0} & \textbf{1.0} & 0.5 & 0.3 & \textbf{0.3} \\
    \hline
  \end{tabular}
\end{table}

Finally, success rates of reaching the exact optimum were discussed. 
In the Sphere benchmark, the exact optimum score was trivially $0$. In our Number Partitioning settings, the exact optimum scores were $1,0,0$ for $D=20, 30, 40$, respectively.
Thus, we evaluated the success rates by calculating the proportion of runs in which the optimized score corresponded to these values.
The results were summarized in Table~\ref{tab:success_rate}.
The proposed method achieved top-tier performance comparable to \texttt{BO-UCB}.

\section{Discussion}

The results demonstrate that discrete kernel surrogates admit an exact and computationally structured optimization procedure through integer linear programming. The central advantage of ILP-BO is not merely the use of an integer programming solver, but the exact reformulation of nonlinear kernel surrogate optimization into a 0--1 ILP. By exploiting the finite set of possible values of discrete distance or similarity measures and introducing binary one-hot variables, the proposed formulation optimizes the surrogate model without approximating the kernel function. Consequently, the candidate-selection problem can be solved to global optimality within the specified solver tolerances, providing a stronger optimization guarantee than heuristic or local optimization procedures commonly used for acquisition optimization.

A second important property of ILP-BO is that exploration can be incorporated without sacrificing this linear structure. The Hamming-distance threshold $l_{\mathrm{th}}$ excludes neighborhoods around previously observed points and therefore imposes a direct geometric form of exploration in the binary search space. At the same time, increasing $l_{\mathrm{th}}$ reduces the number of auxiliary binary variables from $KL$ to $K(L-l_{\mathrm{th}})$. Thus, exploration and optimization complexity are coupled in a favorable way: stronger exclusion of previously explored regions simultaneously reduces the size of the resulting ILP. This property distinguishes the proposed framework from uncertainty-based exploration mechanisms whose optimization can introduce additional nonlinear structure.

The decay-scale analysis further clarifies the role of this exploration mechanism. Smaller values of decay scales $r$ rapidly reduce the Hamming-distance margin and thereby favor exploitation at earlier iterations, whereas larger values place greater emphasis on exploration in later iterations. The observed trade-off between early and late optimization progress supports the use of an intermediate decay scale for the single-scale schedule and motivates the multiple-scale schedule, which combines different exploration rates over successive iterations. Rather than treating the decay parameter as an arbitrary fixed hyperparameter, these results provide an interpretable view of $r$ as a control parameter governing the temporal balance between exploration and exploitation.
The present experiments do not isolate the contribution of exact ILP optimization from those of the surrogate and such Hamming-distance schedule.
Disentangling these individual effects is left for future work.

The proposed framework has an inherent computational limitation. The number of auxiliary binary variables grows as $K(L-l_{\mathrm{th}})$, and therefore generally increases as additional observations are collected and the exploration margin decreases. For long optimization horizons, the cost of repeatedly solving the resulting ILPs may become substantial relative to lightweight heuristic acquisition optimizers. This limitation is particularly relevant when the objective function itself is inexpensive to evaluate. In such settings, the additional optimization overhead of ILP-BO may exceed the reduction in black-box evaluations.

Several extensions could mitigate this limitation. A two-sided Hamming-distance band, $l_{\min} \le l \le l_{\max}$, would restrict the candidate to a finite shell around the current observations and reduce the number of auxiliary variables to $K(\Delta l+1)$, where $\Delta l=l_{\max}-l_{\min}$. This formulation can substantially reduce the dependence of candidate-selection complexity on the full problem dimension. Another possibility is to switch from ILP optimization to brute-force enumeration when the search margin becomes sufficiently small. For example, when candidates are restricted to Hamming distance one, the number of distinct neighboring candidates generated from the observed points is at most $\mathcal{O}(KD)$. Such hybrid strategies may provide an effective transition from globally optimized ILP-based candidate selection to lightweight local enumeration search in the late stages of optimization.
Moreover, to keep the number of training data $K$ constant, one can consider removing unimportant data.
Constructing a surrogate model via support vector machines could also be effective in aggressive reduction of $K$.

Importantly, the practical value of this computational trade-off depends on the cost of the black-box evaluation. In expensive applications such as molecular design, materials discovery, experimental process optimization, or hardware configuration, each objective evaluation may require substantial physical or computational resources. In such sample-limited settings, the ability to identify high-quality candidates with relatively few black-box evaluations can outweigh the computational overhead of the surrogate optimization itself. The experimental results support this perspective: while the relative performance of ILP-BO depends on the benchmark and dimensionality, the proposed method largely exhibits competitive early-stage behavior, and in several discrete optimization problems it attains the best or near-best solutions within the limited evaluation budget.

Overall, the results suggest that the principal contribution of ILP-BO is a general optimization principle rather than a universally dominant benchmark-specific heuristic. The combination of exact discrete kernel linearization, structured Hamming-distance exploration, and globally optimized candidate selection provides a mathematically transparent framework for discrete black-box optimization. This perspective opens the possibility of extending the same formulation to broader classes of discrete surrogate models and search spaces.

\section{Conclusion}

We introduced ILP-BO, an integer linear programming-based framework for discrete black-box optimization. The key idea is to exploit the finite-valued structure of discrete distance or similarity measures to transform nonlinear kernel surrogate optimization into an exact 0--1 integer linear program using binary one-hot auxiliary variables. This formulation eliminates the need for kernel approximation during candidate selection and enables the surrogate optimization problem to be solved to global optimality within solver tolerances.

ILP-BO further incorporates exploration through a Hamming-distance margin around previously observed solutions. This mechanism provides an interpretable control of the exploration--exploitation trade-off while simultaneously reducing the number of auxiliary variables in the ILP. We additionally derived an analytical upper bound on the Hamming-distance threshold from the coverage of the binary search space, providing a principled criterion for maintaining feasibility.

Experiments on Sphere, Max-Cut, Number Partitioning, Ising, and LABS benchmarks demonstrated that the proposed framework is competitive with established discrete Bayesian optimization methods and can provide strong sample efficiency. The decay-scale study further revealed a systematic trade-off between early exploitation and sustained exploration, motivating the proposed multiple-scale schedule. Moreover, the benchmark experiments showed that ILP-BO can attain known global optima for several problem settings within the limited evaluation budget.

The results establish a general framework for exact surrogate optimization in discrete black-box problems rather than a universally superior optimizer for every benchmark. Future work will investigate extensions to general integer and mixed discrete domains, adaptive and two-sided trust-region mechanisms, training-set compression, and hybrid candidate-selection strategies that combine integer programming with direct enumeration. These extensions may enable ILP-BO to support longer optimization horizons and larger-scale discrete search problems while retaining its central advantage of exact surrogate optimization.

\subsubsection*{Broader Impact Statement}
This work presents a general algorithmic framework for discrete optimization, which positively impacts sample efficiency in expensive scientific and engineering tasks without introducing direct negative societal risks.

By converting discrete kernel surrogate optimization into an integer linear program, ILP-BO enables the next candidate to be selected through a globally optimized mathematical-programming problem rather than a heuristic procedure.
This allows us to more clearly identify whether the issue lies in the surrogate model itself or in the optimization component. In other words, this methodology has the potential to serve as a highly transparent black-box optimization framework.

The potential societal impact is primarily positive because the method targets settings in which each objective evaluation can be expensive, including molecular and drug design, materials discovery, experimental process optimization, and automated hardware configuration. In such applications, reducing the number of required experiments or simulations may substantially lower computational, experimental, and material costs.

\subsubsection*{Acknowledgments}
This work was partially supported by the Council for Science, Technology, and Innovation (CSTI) through the Cross-ministerial Strategic Innovation Promotion Program (SIP), ``Promoting the application of advanced quantum technology platforms to social issues'' (Funding agency: QST), the Japan Society for the Promotion of Science (JSPS) KAKENHI (Grant Number JP23H05447), Japan Science and Technology Agency (JST) (Grant Number JPMJPF2221). 
S. Tanaka wishes to express their gratitude to the World Premier International Research Center Initiative (WPI), MEXT, Japan, for their support of the Human Biology-Microbiome-Quantum Research Center (Bio2Q).

\bibliography{tmlr}
\bibliographystyle{tmlr}

\end{document}